\documentclass[journal,twoside]{IEEEtran}
\usepackage{cite}
\usepackage{amsmath,amssymb,amsfonts}
\usepackage{algorithmic}
\usepackage{graphicx}
\usepackage{booktabs}
\usepackage{adjustbox}
\usepackage{algorithm}
\usepackage{mwe}
\usepackage{subfigure}
\usepackage{subcaption}
\usepackage{booktabs}
\usepackage{multirow}
\usepackage{booktabs}
\usepackage{array}
\usepackage{adjustbox}
\usepackage{soul}
\usepackage{float}
\usepackage{array}
\usepackage{subfigure} 
\usepackage{textcomp}

\def\BibTeX{{\rm B\kern-.05em{\sc i\kern-.025em b}\kern-.08em
    T\kern-.1667em\lower.7ex\hbox{E}\kern-.125emX}}

\begin{document}
\title{
RL-ACRGNet: Reinforcement Learning-Based Chest Radiology Report Generation Network}
\author{Yogesh Kumar Meena, Saurabh Agarwal, and K.V. Arya
\thanks{Yogesh Kumar Meena is with Human-AI Interaction (HAIx) Lab, Indian Institute of Technology Gandhinagar, India (e-mail: yk.meena@iitgn.ac.in).}
\thanks{Saurabh Agarwal is with the Department of Computer Science and Engineering, Madhav Institute of Technology and Science Deemed University (MITS-DU), Gwalior, India.}
\thanks{K. V. Arya is with the Multimedia and Information Security Research Group, Department of Computer Science and Engineering, ABV-Indian Institute of Information Technology and
Management, Gwalior 474015, India.}}

 \maketitle
\begin{abstract}
Medical imaging interpretation is a foundational pillar of modern clinical diagnostics, yet the manual generation of radiology reports remains a time-consuming process prone to interpretation inconsistencies. Within the field of medical AI, automating these descriptions through deep learning promises to streamline clinical workflows and standardise diagnostic output. However, accurate disease detection and precise report generation remain significant challenges due to limitations in capturing fine-grained visual features and ensuring clinical coherence. To address these issues, we propose RL-ACRGNet, an improved encoder-decoder model that integrates a pre-trained DenseNet encoder with a multilevel LSTM decoder within an off-policy reinforcement learning framework. Using a dual-network approach to refine visual-semantic embeddings through a metric-based reward mechanism, we demonstrate that RL-ACRGNet consistently outperforms state-of-the-art baselines on the IU-Xray dataset, achieving quantitative improvements in BLEU-4 (0.47\%), METEOR (0.17\%) and ROUGE-L (0.518). Furthermore, comprehensive evaluations on the large-scale MIMIC-CXR data set confirm the robust generalisation of the model and its ability to generate high-quality, clinically relevant reports. 
\end{abstract}


\section{Introduction}
Pulmonary diseases represent a significant threat to public health, with manifestations that can span from the oral cavity to the lungs~\cite{barnes2003chronic}. Detecting and diagnosing these conditions hinges critically on advanced medical imaging techniques, such as chest X-rays (CXR) and computed tomography (CT) scans. However, the complexity of these images requires specialised knowledge for accurate interpretation, traditionally relying on the expertise of trained radiologists~\cite{saporta2022benchmarking}. Recently, automatic chest X-ray report generation systems (ACRGS) \cite{acrg1,acrg2,acrg3} have emerged as a transformative solution in this domain, enabling automated analysis of medical images and the generation of detailed reports that closely mimic a radiologist's assessments. 

Chest X-ray (CXR) images play a significant role in visualising key structures within the chest, including the lungs and heart. To ensure clear and comprehensive radiology reports, careful interpretation of these images is essential. The ABCED mnemonic \cite{ABCDE} is a valuable tool for systematically evaluating CXR images, helping to ensure accurate reporting of critical findings. 

Additionally, advanced encoder-decoder-based image reporting models \cite{alfarghaly27automated, bwang2022automated, yu2025, wang2025} have gained popularity for generating concise and accurate radiology reports. These models combine deep learning (DL) techniques \cite{McCulloch}, reinforcement learning (RL), and hierarchical modelling. convolutional neural networks (CNNs) [\citenum{Fukushima}] have been widely adopted as a fundamental component in image reporting systems. It is utilised as an encoder in the ACRGNet model. CNNs excel at extracting high-level visual features from images, enabling them to build an understanding of the content and context of visual input. The pretrained DenseNet model \cite{huang2017densely} was utilised as a CNN (encoder) to extract deep features from CXR images. The DenseNet has architectural diversity and comprises a heterogeneous mix of convolutional layers with various kernel sizes (ranging from 1$\times$1 to 7$\times7$) and grouping configurations (ranging from ungrouped to depth-wise). This diversity underscores the meticulous tailoring of the network architecture at the granular level for detecting pulmonary disease from CXR images. Hierarchical modelling using long short-term memory (LSTM) [\citenum{Rumelhart}] has proven effective for generating reports by capturing temporal word dependencies and maintaining context throughout the report generation process. By leveraging the hierarchical nature of language, LSTM facilitates the generation of coherent, contextually relevant reports.

Recent advancements in image reporting have integrated reinforcement learning techniques to address the limitations of earlier CNN- and LSTM-based approaches in capturing detailed context and global coherence. Reinforcement learning, as outlined by [\citenum{Sutton}] and implemented in studies such as [\citenum{Ren}], introduces an agent-environment interaction framework in which neural networks serve as agents, generating reports for images. Each image acts as a state within the environment, guiding the agent's actions in generating sequential reports. Through trial and error, the agent refines its reporting strategy by receiving rewards based on the quality of its generated reports. A significant advancement in this field is adopting the actor-critic approach, which combines policy, value, and reward networks. Considering the prior context, the policy network directs word generation locally, while the value network evaluates global coherence across potential sentence completions. Meanwhile, the reward network scores partially generated reports, thereby crucially informing the training processes of both the policy and value networks. This integrated approach has demonstrated substantial improvements in the effectiveness of image reporting systems, underscoring its potential to advance future research in the field. 

Despite the progress made in this area, challenges remain in achieving semantic consistency between medical images and their textual representations. Advanced frameworks, such as multi-grained report generation \cite{liu2024multi} and hybrid-reward-based reinforcement learning \cite{y, k}, have been proposed to enhance the alignment of image-sentence pairs without requiring extensive annotations. These ongoing research efforts highlight the potential for ACRGS to streamline medical workflows and improve diagnostic accuracy. 

This work introduces a comprehensive framework for ACRGS that integrates a DenseNet encoder with a multilevel LSTM decoder via an off-policy actor-critic reinforcement learning approach. By addressing the limitations of traditional sequence-to-sequence models, this hybrid strategy yields more accurate, diverse, and contextually grounded clinical narratives. This paper significantly extends the diagnostic foundations laid in our previous works—CNN-O-ELMNet \cite{agarwal2024cnn}, MultiFusionNet \cite{agarwal2024multifusionnet}, and CXR-Net \cite{agarwalcxrnet}—by evolving from automated pulmonary disease diagnosis to full-scale descriptive reporting for chest X-ray (CXR) images. The major contributions of this work are as follows:

\begin{enumerate}
   
\item Proposing RL-ACRGNet, a hybrid reinforcement learning–based automatic report generation network for CXR images that streamlines image analysis and clinical reporting, improving radiologists' efficiency and accelerating diagnostic decisions. 

\item Designing a unified architecture pairing a DenseNet visual encoder with a multilevel LSTM decoder, enabling the model to effectively capture both fine-grained visual patterns and long-term linguistic dependencies. 

\item Developing an enhanced off-policy actor–critic strategy utilising dual networks for word-level refinement and sequence-level coherence. By employing a composite reward function that linearly combines standardised NLP metrics, the model directly optimises for clinical accuracy and report diversity, significantly outperforming current state-of-the-art methods on benchmark datasets.

\end{enumerate}

The remaining paper is organised as follows: \textbf{Section II} describes key related work in radiology report generation, including hand-crafted features, deep learning-based approaches, encoder-decoder frameworks, and reinforcement learning-based research.
\textbf{Section III} explains the entire method, including problem formation, description of all three networks, the training procedure, and the inference mechanism.
\textbf{Section IV} is all about the experimental process, including data preparation, network implementations, hyperparameters, and results with discussion. Finally, \textbf{Section V} presents the paper's conclusions, provides a short summary of the proposed approach, and offers suggestions for future expansion of the scope. 

\section{Related Work}

Recent advances in deep learning have significantly improved automated generation of medical reports for pulmonary diseases. This section reviews various approaches aimed at enhancing report accuracy~\cite{ cpandey2022comprehensive, dliao2023deep, esalahuddin2022transparency, fJing_2018, van2022explainable,singh2022efficient, gajbhiye2022translating}. A key method is the CNN-RNN framework, which employs Convolutional Neural Networks (CNNs) to extract visual features and Recurrent Neural Networks (RNNs) to generate detailed reports, marking a significant progress in automated reporting systems.

\textbf{CNN-RNN-based Framework}

The innovative research detailed in \cite{fJing_2018} on radiology report generation introduces a novel approach by integrating co-attention mechanisms with HRNN networks. This model detects medical tags, derives visual features through a CNN's multi-label classification process on chest X-rays (CXR), and generates sentences based on the anticipated tags using a combination of co-attention and hierarchical LSTM techniques. Although limitations include a limited vocabulary set, which can lead to potential bias. TieNet \cite{wang2018tienet6} employs a simple CNN-RNN architecture for categorising common thoracic diseases and incorporates spatial attention for preliminary X-ray report generation. In \cite{12yu2018topic}, Yu et al. employed order embedding to describe topic-oriented images and utilised a CNN-based classifier to choose topics for images from a pool of candidates. A recurrent memory network (RMN) is developed by Wang et al. \cite{13wang2020retrieval} to report images. During training, topic words were recorded from a topic repository. Testing involved a retrieval method to generate a topic word, which was then incorporated into a sentence through a recurrent memory network. Capturing abstract concepts or relationships among image elements can be challenging for these models, leading to superficial or incomplete reports.

Zhang et al. \cite{3zhang2019vaa} utilised the visual aligning attention DCNet for image reporting. They developed a visual alignment loss function to enhance the attention layer during training by using a visual vocabulary to exclude non-visual words and reduce their influence on the attention mechanism. Zhou et al. \cite{14zhou2019re} introduced a saliency-enhanced re-reporting model that first extracts semantic and visual saliency cues and then integrates these cues to self-boost the model. In \cite{15zhao2020cross}, Zhao et al. utilised DCNet adaptation and cross-modal retrieval techniques for image reporting across different domains. This approach involves pre-training the cross-modal model on the source domain and fine-tuning image-sentence pairs through a retrieval model. However, these models may struggle to capture detailed contextual information in images, leading to generic or less accurate reports.

Cross-domain image reporting using machine learning was implemented by Yang et al. \cite{16yang2018multitask}, where a CNN-LSTM generated textual explanations for images, and a conditional generative adversarial network synthesised images from text descriptions to generate accurate CXR reports. Hoxha et al. \cite{16yang2018multitask} used CNNs and RNNs for image reporting, extracting visual features that were translated into textual explanations. Similar images were retrieved by comparing vectors of textual explanations with archive images. In \cite{24xian2019self},  Xian et al. developed an image reporting model based on multimodal LSTM. These models can be computationally intensive, limiting their scalability for large-scale deployment in real-world applications.

\textbf{CNN-RNN-RL-based Framework}
The HRGR model, which combines retrieval and generation via reinforcement learning (RL) \cite{li2017deep}, introduces a hierarchical decision-making approach to decide whether to extract sentences from an existing dataset or generate new descriptions for X-ray images. Meanwhile, a domain-focused hierarchical CNN-RNN model integrated with RL \cite{liu9clinically} prioritises clinical precision by using RL rewards to enhance the CIDEr score. Other approaches, including cooperative multi-agent systems, adversarial reinforcement learning, and discriminators for language fluency and diagnostic accuracy, contribute to RL-based frameworks.

\textbf{Miscellaneous Framework}
The Transformer architecture, which utilises multiple layers of multi-head attention, proves highly effective for natural language processing tasks such as generating medical reports. A model for medical imaging reports based on the Transformer, as detailed in \cite{xiong24reinforced}, achieves strong performance in optimising CIDEr rewards through reinforcement learning. Different transformer variants, such as two-way encoders and decoders, relational memory, and conditioned Transformers, showcase innovations for handling medical imaging data and addressing challenges in report generation. 

Yang et al. \cite{19yang2020ensemble} developed an image reporting model that leverages a dual-gan generator and utilises generation- and retrieval-based methods. Yuan et al. \cite{20yuan2019exploring} suggested a method for generating image reports using a graph convolution network and multi-level attention for focusing on attributes. While the attribute graph convolution module learns the required attribute features for reporting, this method employs an attention mechanism to highlight specific spatial and scale properties.
Huang et al. \cite{21huang2020denoising} introduced a multi-scale feature fusion technique for image reporting using a denoising approach. Additionally, Monay et al. \cite{23yu2019multimodal} developed a model that equally incorporates textual and visual data through an expectation-maximisation algorithm. Additionally, Yu et al. developed a DCNet for image reporting based on a multimodal transformer model, which incorporates multi-view visual features to enhance performance. Despite their advancements, these methods are computationally intensive, demanding substantial resources and time for both training and inference, which could hinder their feasibility in real-time or resource-limited situations. Medical Image Captioning using CvT and DistillGPT2 generates CXR reports to reduce radiologist workload using convolution vision transformer in \cite{2024a}, while ChestX-Transcribe \cite{2025a} uses Swin Transformer and DistilGPT for high-resolution feature extraction and report generation, outperforming baselines in BLEU/ROUGE scores.

The proposed approach addresses existing model limitations by introducing a CNN-RNN-RL-based model that provides flexibility for learning inter-modality patterns and generating comprehensive radiology reports across various CXR views. The subsequent sections discuss the proposed methods and their outcomes.

 \section{Proposed method}
The proposed model, RL-ACRGNet, is designed to generate radiology reports/captions for lung diseases. It is a hybrid model that incorporates three cutting-edge deep learning technologies: CNNs, RNNs, and RL. In the RL-ACRGNet model, the following terminology is utilised:
\begin{itemize}
\item The Agent encompasses both the policy network $q_\pi$ and the value network $V_\theta$.
\item The Environment comprises both the input image $I$ and the subsequent words predicted from $(s_1, s_2, ..., s_t)$, where $s_i$ represents the $i^{th}$ word in the sentence $S$.
\item The Action $(b)$ involves predicting the next word $s_{t+1}$ from the dictionary of words $Y$.
\item The State $(r)$ consists of the image $I$ and the predicted words.
\end{itemize}

In this approach, the agent interacts with the environment by taking actions to achieve a specific goal. The primary goal here is to generate a sentence 
S = ($s_1, s_2, ..., s_t)$ that precisely captures the details depicted in an image I. This section comprehensively explains the method for generating radiology reports using reinforcement learning. It delves into the architecture of the involved networks, including the CNN-RNN, policy, value, and reward networks. Furthermore, it describes the process for training these models.

\subsection{CNN-RNN Network}
\begin{figure}
    \centering
    \includegraphics[width = 3.4in, height= 4in]{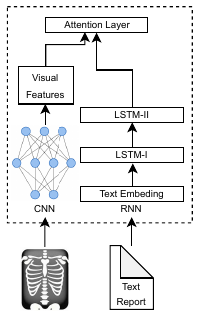}
    \caption{Architecture of CNN-RNN network}
    \label{fig:cnn-rnn}
\end{figure}

The proposed model combines CNNs, RNNs, and attention mechanisms to generate descriptive reports for medical images. 
Fig.~\ref{fig:cnn-rnn} shows the CNN-RNN networks' architecture, which processes related textual reports using multi-level Long Short-Term Memory (LSTM) layers and extracts detailed features from CXR images using a pre-trained DenseNet CNN.
A multihead-attention mechanism enhances the relevance and coherence of generated reports by selectively emphasising significant words in the input sequence. The model begins by tokenising the text and embedding each token using pre-trained word embeddings, capturing semantic relationships effectively. Two LSTM layers, LSTM-I with 256 units and LSTM-II with 512 units, equipped with 'tanh' activation functions, sequentially capture temporal dependencies and hierarchical features from the textual input, ensuring comprehensive report generation for medical images.

To enhance the model's ability to focus on pertinent parts of both the input sequence and the CXR image, an attention mechanism is integrated after the LSTM-II layer. This mechanism combines the visual and textual information to generate meaningful reports. Subsequently, dense layers with dropout regularisation are employed to transform the combined features into the vocabulary space, facilitating report generation.  

\begin{figure}[ht]
    \centering
    \includegraphics[width = 3.4in]{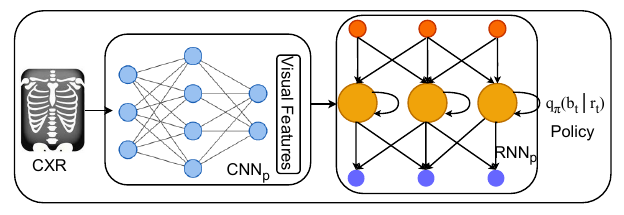}
    \caption{Representing the structure of the policy network, including CNN as an encoder and RNN as a decoder}
    \label{fig:policy}
\end{figure}

\subsection{Policy Network}

The policy network is designed to assign actions to specific states. It calculates the probability of each action based on the current state. The structure of the policy network is shown in  Fig. \ref{fig:policy}. This policy network, denoted as $q_\pi$, is utilized to estimate the policy $q_\pi(b_t|r_t)$, where the current state is represented as $r_t = \{ \textbf{\textit{I}}, a_1, a_2, ... , a_t \}$ and the action is $b_t = a_{t+1}$. The policy network architecture comprises a CNN and an RNN, resembling the encoder-decoder architecture commonly used in various image reporting models. These components are referred to as CNNq and RNNq, respectively. CNNq processes the visual data from the CXR images and encodes them. The encoded data is then fed as input to the initial node $i_0$ of RNNq. When an action $b_t$ is taken at each time step, the policy is represented by the RNNq's hidden state $z_t$, which evolves over time. As each action is taken, the state transitions from $z_t$ to $z_{t+1}$, reintroducing the generated word into the input of RNNq. The network's operation can be summarised by the following set of equations.

\begin{equation}
    i_0 = M^{i,v}CNN_q(\textit{I})
\end{equation}
\begin{equation}
    z_t = RNN_q(z_{t-1},i_t)
\end{equation}
\begin{equation}
    i_t = \varphi(s_{t-1}),  t>0
\end{equation}
\begin{equation}
    \pi(b_t|r_t) = \phi(z_t)
\end{equation}

Where $\phi$ and $\varphi$ are the output and input models of RNN$_q$ respectively, and $M^{i,v}$ is the weight matrix of the linear embedding model of the visual information.

\subsection{Value Network}
Value network $V_\theta$ is based on the value function $Vq$ of a policy $q_\pi$. The value function $Vq$ is defined as an expectation over the total reward $R$ under a policy $q_\pi$:
\begin{equation}
    V_q(b) = E[R|r_t, b_{t \dots T} \sim q_\pi ]
\end{equation}
Here, $r$ represents the state, and $b_{t \dots T}$ represents a sequence of actions taken from time step $t$ to $T$, as per the policy network $q_\pi$.

\begin{figure}
    \centering
    \includegraphics[width = 3.4in]{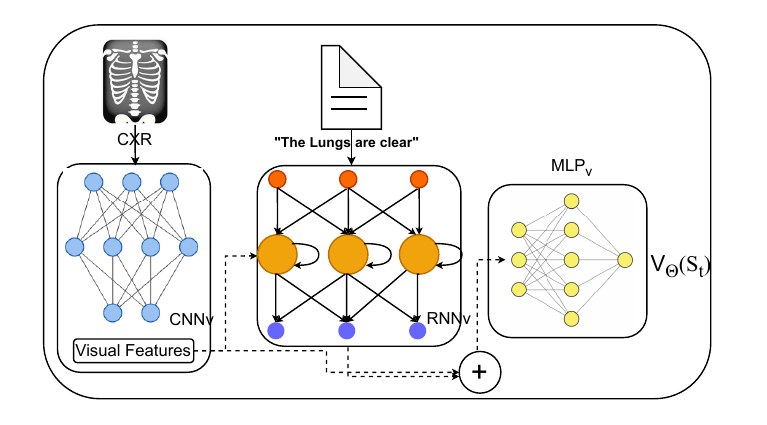}
    \caption{Diagram representing the structure of the value network}
    \label{fig:value}
\end{figure}

The value network approximates this value function. The value network structure is shown in Fig. \ref{fig:value}. It starts with a CNN for extracting visual features of the image and consists of an RNN that encodes the linguistic information of the incompletely predicted sequence $(s_1, s_2, ..., s_t)$. Then, both these pieces of information are connected using a Multi-Layer Perceptron (MLP). It predicts the expected return for the sequence by accounting for the reward associated with the report. These are denoted as CNN$_p$, RNN$_p$, and MLP$_v$, respectively.

\subsection{Reward Network}
\begin{figure}
    \centering
    \includegraphics[width = 3.4in]{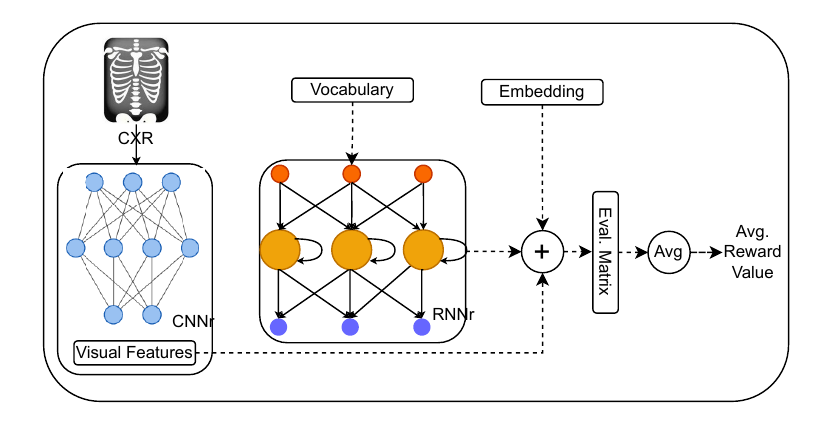}
    \caption{Reward Network architecture}
    \label{fig:reward}
\end{figure}

The reward network is responsible for generating the reward for the generated medical image caption. It gives the rewards in terms of score which is supposed to be maximised. Fig.~\ref{fig:reward} shows the structure of the reward network, which contains a CNN and an RNN denoted as CNN$_r$ and RNN$_r$. It included a mapping linear layer represented as $lm_l$. The last hidden state of RNN$_r$ for a sentence C corresponds to the embedding feature $hs^{'}{_T}(C)$. The feature vector extracted by CNN$_r$ of the image $F$ is denoted as $k$. $lm_l(.)$ is the function for mapping features of the image to the embedding space. This network is trained on the same image-sentence pairs as in other networks. The CNN$_r$ is not trainable, but the rest is. RNN$_r$ and $lm_l(.)$ are trained using a bidirectional ranking loss, which is defined as follows:

\begin{equation}
\begin{aligned}
BRL_r & =\sum_{\boldsymbol{k}} \sum_{C^{-}} \max \left(0, \gamma-lm_l(\boldsymbol{k}) \cdot \boldsymbol{hs}_T^{\prime}(C) \right.\\ 
& \left. +lm_l(\boldsymbol{k}) \cdot \boldsymbol{hs}_T^{\prime}\left(C^{-}\right)\right) \\
& +\sum_C \sum_{\boldsymbol{k}^{-}} \max \left(0, \gamma-\boldsymbol{hs}_T^{\prime}(C) \cdot lm_l(\boldsymbol{k}) \right.\\
& \left. +\boldsymbol{hs}_T^{\prime}(C) \cdot lm_l\left(\boldsymbol{k}^{-}\right)\right)
\end{aligned}
\label{rewloss}
\end{equation}

where $\gamma$ is the cross-validation margin, $(k, C)$ are true image-report pairs, $C^-$ represents the negative description of the image feature $k$ and vice versa for $k^-$.

For image features $k^*$, the reward for the predicted report $\widehat{C}$ is the normalised distance between $\widehat{C}$ and $k^*$ :
\begin{equation}
r_1=\frac{lm_l\left(\boldsymbol{k}^*\right) \cdot \boldsymbol{hs}_T^{\prime}(\widehat{C})}{\left\|lm_l\left(\boldsymbol{k}^*\right)\right\|\left\|\boldsymbol{hs}_T^{\prime}(\widehat{C})\right\|}
\end{equation}

The second part of the reward, $r_2$ is given by:
\begin{equation}
\begin{aligned}
    r_2 & = 1/6 [e_1 (BLEU1) + e_2 (BLEU2) + e_3 (BLEU3)  
    \\ 
    & + e_4 (BLEU4) + e_5 (ROUGE-L) + e_6 (METEOR)]
\end{aligned}
\end{equation}

where, $e_1$, $e_2$, ... , $e_7$ are hyperparameter weights for each evaluation metric. These weights are to be tuned during experimentation.

The final reward used for training is 
\begin{equation}
    r=r_1 + r_2
\end{equation}

\subsection{Training}
The network is trained in two steps. In the first step, all three networks are trained individually. The reward network is trained using the loss described in Eq. \ref{rewloss}. Cross entropy loss, given in Eq. \ref{cross-entropy-loss}, is used to train the policy network.  Loss is calculated in the value network using Mean Squared Error $\left\|u_\theta\left(r_i\right)-o\right\|^2$, for training purposes. 

\begin{equation}
    CE_{q^{\prime}}=-\log q\left(s_1, \ldots, s_T \mid \mathbf{I} ; q_\pi\right)=-\sum_{t=1}^T \log q_\pi\left(b_t \mid r_t\right)
    \label{cross-entropy-loss}
\end{equation}

After the first phase of training involving the policy and value networks, the off-policy algorithm is applied, along with curriculum learning. This algorithm improves the primary policy, which is the optimal policy to learn while maintaining a separate policy for taking actions in the environment. Curriculum learning is used to handle the large action space by training the actor-critic model with sentences of increasing difficulty. Initially, the last few words of the sentence are used for actor-critic training, gradually increasing until the entire sentence is used. 

\subsection{Inference}
Inference is a crucial step in predicting image descriptions using deep learning models. It involves using a trained model to predict a sentence that corresponds to a given image. In this case, inference involves using beam search (with beam size B) along with the value and policy networks.

The start token and the image features are the first steps in the inference process. The policy probabilities for every word to be created are then determined. Given the image and a partially predicted sentence, the policy network generates the probability of the subsequent word.
Once the policy probabilities have been calculated, the best B candidates are selected for the next word. The extension scores for each candidate are then calculated using the value network. It is used to estimate the quality of the partial report generated so far.
The best B candidates are kept to generate the next word until a candidate sentence ends or the maximum word limit is reached. This process is repeated till an end token is predicted, indicating that the model has generated a complete report for the image.

The beam search algorithm helps keep track of multiple candidate sentences simultaneously, and the value network selects the best candidates at each step. This approach helps to predict more accurate and relevant sentences for the images.

\section{Results and Discussion}

The proposed model is evaluated on two benchmark datasets of CXR images and associated text reports. The following section provides details on the datasets, results, and visual analysis.

\begin{table*}[]
\centering
\caption{Comparison of Medical Image reporting Methods}
\label{tab:image_captioning}
\begin{tabular}{@{}clcccccc@{}}
\toprule
\textbf{Dataset} & \multicolumn{1}{c}{\textbf{Method}} & \textbf{BLEU-1} & \textbf{BLEU-2} & \textbf{BLEU-3} & \textbf{BLEU-4} & \textbf{METEOR} & \textbf{ROUGE-L} \\ \midrule
\multirow{8}{*}{IU-Xray} &
  AdaAttn~\cite{50lu2017knowing} &
  0.478 &
  0.314 &
  0.224 &
  0.166 &
  0.203 &
  0.390 \\
 &
  METransformer~\cite{wang2023metransformer} &
  0.483 &
  0.322 &
  0.228 &
  0.172 &
  0.192 &
  0.380 \\
 &
  R2Gen~\cite{35chen2020generating} &
  0.429 &
  0.307 &
  0.232 &
  0.179 &
  0.195 &
  0.426 \\
 &
  DenseNet with RL~\cite{NEURIPS2018_e0741335} &
  0.438 &
  0.298 &
  0.208 &
  0.151 &
  0.322 &
  0.414 \\
 &
  PPKED~\cite{51liu2021exploring} &
  0.491 &
  0.321 &
  0.241 &
  0.155 &
  0.149 &
  0.377 \\
 &
  PureT~\cite{wang2022automated} &
  0.496 &
  0.319 &
  0.241 &
  0.175 &
  - &
  .0377 \\
 &
  KiUT~\cite{huang2023kiut} &
  0.525 &
  0.360 &
  0.251 &
  0.185 &
  0.242 &
  0.409 \\
 &
  \textbf{Proposed Method} &
  \textbf{0.592} &
  \textbf{0.447} &
  \textbf{0.387} &
  \textbf{0.276} &
  \textbf{0.384} &
  \textbf{0.518} \\ \midrule
\multirow{8}{*}{MIMIC-CXR} &
  AdaAttn~\cite{50lu2017knowing} &
  0.299 &
  0.185 &
  0.124 &
  0.088 &
  0.118 &
  0.266 \\
 &
  METransformer~\cite{wang2023metransformer} &
  0.386 &
  0.250 &
  0.169 &
  0.124 &
  0.152 &
  0.291 \\
 &
  R2Gen~\cite{35chen2020generating} &
  0.352 &
  0.218 &
  0.145 &
  0.103 &
  0.142 &
  0.277 \\
 &
  DenseNet with RL~\cite{NEURIPS2018_e0741335} &
  0.348 &
  0.237 &
  0.135 &
  0.114 &
  0.146 &
  0.268 \\
 &
  PPKED~\cite{51liu2021exploring} &
  0.360 &
  0.224 &
  0.149 &
  0.106 &
  0.149 &
  0.284 \\
 &
  PureT~\cite{wang2022automated} &
  0.351 &
  0.223 &
  0.157 &
  0.118 &
  - &
  0.287 \\
 &
  KiUT~\cite{huang2023kiut} &
  0.393 &
  0.243 &
  0.159 &
  0.113 &
  0.160 &
  0.285 \\
 &
  \textbf{Proposed Method} &
  \textbf{0.401} &
  \textbf{0.258} &
  \textbf{0.171} &
  \textbf{0.124} &
  \textbf{0.164} &
  \textbf{0.274} \\ \bottomrule
\end{tabular}
\end{table*}

\begin{figure*}
    \centering
    \includegraphics[width = 7in]{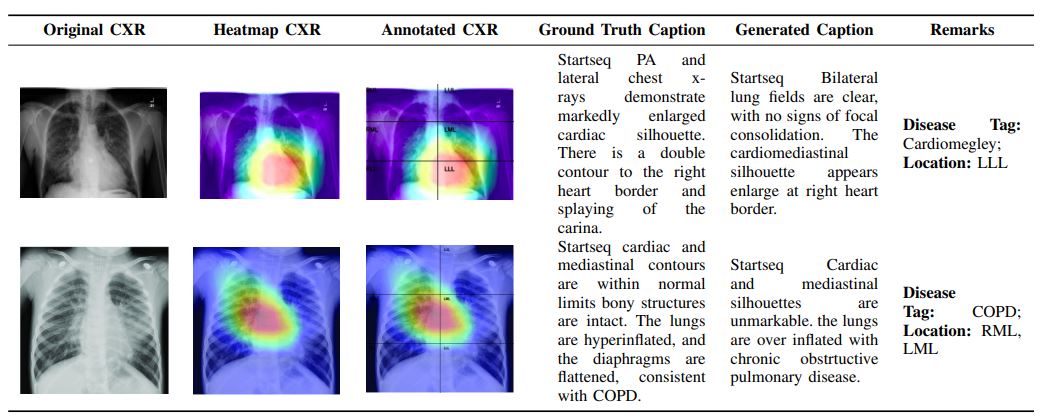}
    \caption{Qualitative analysis of the proposed model demonstrates its effectiveness in highlighting diseased areas, providing detailed annotations regarding the disease's location and relevant diagnostic tags.}
    \label{fig:visual results}
\end{figure*}

\subsection{Dataset and Evaluation Metrics}
Multiple experiments are performed on the benchmark datasets IU-Xray~\cite{demner2016preparing} and MIMIC-CXR~\cite{johnson2019mimic}. To maintain consistency and fairness, we meticulously adhered to the dataset splitting and preprocessing methodologies outlined by Nguyen et al.~\cite{nguyen2021automated}.

The IU-Xray dataset has 7470 CXR images in lateral and PA views. These images are accompanied by 3955 radiology reports provided in XML format. The Indiana University Hospital networks created the dataset, which is publicly accessible. Similarly, the MIMIC-CXR dataset comprises 473,057 CXR images. This dataset includes 206,563 detailed reports and is available for research and development. Both datasets provide valuable resources for studying and advancing medical image analysis.

The performance of the proposed system is evaluated using widely accepted evaluation metrics, such as BLEU \cite{bleu}, CIDEr \cite{cider}, ROUGE \cite{rouge}, and METEOR \cite{meteor}.

\subsection{Evaluation on different metrics}
The quantitative results are presented in Table\ref{tab:image_captioning}. These BLEU scores highlight the model's ability to generate more accurate, contextually relevant single words and longer word sequences, thereby improving the linguistic quality of the generated reports. Additionally, the model substantially improved by 0.518 and 0.274 in ROUGE-L, indicating enhanced recall and precision in capturing the overlap between evaluated and actual reports. The scores of 0.384 and 0.164 in METEOR further emphasise the model's proficiency in producing reports that align more closely with human-perceived quality, taking into account both linguistic and content-based metrics. Overall, these significant improvements across a range of metrics substantiate the effectiveness of the proposed model in advancing the quality of image reports. 

The proposed model is compared with the current state-of-the-art (SOTA) methods including AdaAttn~\cite{50lu2017knowing}, METransformer~\cite{wang2023metransformer}, R2Gen~\cite{35chen2020generating}, DenseNet with RL~\cite{NEURIPS2018_e0741335}, PPKED~\cite{51liu2021exploring}, PureT~\cite{wang2022automated}, and KiUT~\cite{huang2023kiut} and reported in Table \ref{tab:image_captioning}. It is clearly visible that the proposed network outperformed all the SOTA methods at all the evaluation metrics used in this experiment and exhibited an improvement of 0.27\% in BLEU-1, 0.35\% in BLEU-2, 0.48\% in BLEU-3, 0.47\% in BLEU-4, and 0.17\% in METEOR over the baseline reinforcement learning model \cite{NEURIPS2018_e0741335} in IU-Xray dataset. Although the proposed model outperformed all SOTA methods on the MIMIC-CXR dataset except METransformer, it fell short by 0.017 in the ROUGH-L metric.

\subsection{Visual Analysis}
The comparison between the generated report for the CXR and the ground truth report is reported in Fig.~\ref{fig:visual results}. Along with the predicted report, we also provided additional information about the disease, including its location and tags. We also highlight the visual effect on lungs using Gradient-weighted Class Activation Mapping (Grad-CAM)~\cite{selvaraju2017grad}. This outcome underscores the efficacy of RL-ACRGNet in representing results in heatmap form to provide disease location. After we divide the heatmap CXR into 6 regions:  left upper chest (LUC), left middle chest (LMC), left lower chest (LLC), right upper chest (RUC), right middle chest (RMC), and right lower chest (RLC). This process helps to find the location of the disease. Generated reports and remarks clearly demonstrate the model's proficiency in capturing clinically relevant details, anatomical structures, and pathological findings within the images. A robust alignment between generated and ground truth reports indicates the model's potential utility in assisting healthcare professionals by providing accurate and meaningful descriptions of medical images. Moreover, comparing ground truth and generated reports is crucial for building trust and confidence in the model's application in clinical settings, where precise and reliable image interpretation is imperative for informed decision-making and patient care. 

\subsection{Ablation Studies}
The impact of the various components of the proposed model is shown in Table~\ref{tab:ablation}. The BASE model is a basic CNN-RNN architecture tested on both datasets. The '+' symbol denotes the addition of a new component to the existing model. This comparison analyses BASE, BASE+RL, and models with or without Attention to highlight their significance within the overall architecture. The quantitative and qualitative results clearly demonstrate that incorporating additional components enhances the accuracy of the generated reports.

\setlength{\tabcolsep}{3pt} 
\begin{table*}[]
\centering
\caption{Ablation Studies of the proposed model on datasets IU-Xray and MIMIC-CXR. Here, S.E. is State Embedding, and R-L is ROUGE-L.}
\label{tab:ablation}
\setlength{\tabcolsep}{3pt} 
\begin{tabular}{@{}lcccccccccccc@{}}
\toprule
\multirow{2}{*}{\textbf{Method}} & \multicolumn{6}{c}{\textbf{IU-Xray Dataset}}                & \multicolumn{6}{c}{\textbf{MIMIC-CXR Dataset}} \\ \cmidrule(l){2-13} 
 &
  \textbf{BLEU-1} &
  \textbf{BLEU-2} &
  \textbf{BLEU-3} &
  \textbf{BLEU-4} &
  \textbf{METEOR} &
  \textbf{R-L} &
  \textbf{BLEU-1} &
  \textbf{BLEU-2} &
  \textbf{BLEU-3} &
  \textbf{BLEU-4} &
  \textbf{METEOR} &
  \textbf{R-L} \\ \midrule
BASE                   & 0.326 & 0.197 & 0.187 & 0.114 & 0.247 & \multicolumn{1}{c|}{0.358} & 0.247 & 0.158 & 0.121 & 0.089 & 0.135 & 0.248 \\

BASE+RL (w/o Att.) & 0.589 & 0.458 & 0.398 & 0.287 & 0.389 & \multicolumn{1}{c|}{0.512} & 0.395 & 0.265 & 0.165 & 0.121 & 0.159 & 0.270 \\
\textbf{BASE+RL (with Att.)} &
  \textbf{0.592} &
  \textbf{0.447} &
  \textbf{0.387} &
  \textbf{0.276} &
  \textbf{0.384} &
  \multicolumn{1}{c|}{\textbf{0.518}} &
  \textbf{0.401} &
  \textbf{0.258} &
  \textbf{0.164} &
  \textbf{0.124} &
  \textbf{0.164} &
  \textbf{0.274} \\ \bottomrule
\end{tabular}
\end{table*}

\subsection{Discussion}
The proposed model utilises a DenseNet encoder and a multilevel LSTM decoder to efficiently capture hierarchical features from input CXR images and generate coherent, contextually relevant textual reports. We introduced reinforcement learning as a mechanism to fine-tune the generated reports, allowing the model to learn and improve over time. The architecture consists of three main components: a policy network, a value network, and a reward network. Each of these networks serves a different purpose but works together to train the system. 

Both datasets' images are pre-processed and then passed through the DenseNet pre-trained model using the Keras Applications module. The output is then passed through a global average pooling layer to get the final feature vector of size 1536. The entire feature vector of all images is stored and used at multiple stages of training. Start and end tokens are added to each report. The frequency of each word is calculated, and the top 1004 words are kept, while the rest are replaced by an unknown token. This is to reduce the total number of works considered for the predictions. The words are then encoded as integers, and a data structure is used to decode them, which are kept for further reference. The data is ingested via individual reports, with references to the corresponding image features stored in a separate data structure.

Training is done in two phases. First, all three networks are trained separately, and then, the policy and value networks are jointly trained using a curriculum-based off-policy actor-critic algorithm. All training is done using stochastic gradient descent with back-propagation, optimised using the ADAM optimiser [\citenum{Kingma2014AdamAM}]. 

In the first phase of training, the policy network is trained. During each epoch, a batch is sampled from the dataset, and the captions are converted to input captions and output captions by removing the last and the first word of the caption, respectively. The policy network outputs are computed from the input captions, and the outputs are compared with the target captions to compute the loss. It is trained using cross-entropy loss for about 100,000 epochs with a learning rate of $10^{-4}$ and a batch size of 256. 

Next, the reward network is trained using custom loss functions based on bidirectional ranking. The average loss of both directions is added to get the final loss value. It is trained for about 50,000 epochs with a learning rate of $10^{-6}$ and a batch size of 128. This results in a final average loss value of 0.4.

For the value network which gives the estimated reward of all the extensions of a partially generated sentence a custom reward function is written. The first part of the reward is calculated using the normalised distance between the visual and semantic parts of the reward network. For the second part of the reward, an evaluation-based reward is calculated by adding 0.175 times the Bleu-1 score, 0.075 times the Bleu-2 score, 0.075 times the Bleu-3 score, 0.175 times the Bleu-4 score, 0.2 times the ROUGE-L score, and 0.3 times the METEOR score. The scores are calculated using the ground-truth captions for the respective image feature vector. Both parts of the reward are added to get the final reward value.

Finally, the value network is trained using Mean Squared error loss for about 50,000 epochs with a learning rate of $10^{-5}$ and a batch size of 128. This results in a final average loss value of 0.04.

In the second phase of training, the previously trained policy and value networks are further trained using a joint training mechanism called the off-policy actor-critic algorithm. Curriculum learning is also used to reduce the variance in training. It starts by predicting the last few words of the sentence and continues by increasing the length until the entire sentence is predicted by the networks. In the off-policy algorithm, an older policy network is used to predict the next word and the learning is applied to the current network. The log probability of the executed action is calculated using the on-policy network. Reward and value are also calculated using the respective networks. The difference between the reward and the value is called the advantage. Actor loss is calculated using the log probabilities and the advantage, and the critic loss is calculated using the square of the advantage. The total loss is used to train the networks.

The proposed study integrates RL to improve performance and adaptability in medical report generation across various clinical scenarios. It demonstrates strong generalisation across diverse datasets, highlighting its applicability in real-world healthcare settings. However, potential biases in training data and interpretability challenges are acknowledged, emphasising the necessity for rigorous clinical validation. Ultimately, our research advances automated medical image analysis and reporting, offering potential benefits for healthcare efficiency and accuracy.

\section{Conclusion}

The RL-ACRGNet is a new approach to medical report generation that addresses both language fluency and diagnostic accuracy. The approach uses chest radiograph images, with an encoder that extracts visual features and recognises different medical concepts. These ideas are then incorporated into a hierarchical decoder at the sentence and word levels to generate detailed reports.  A specialised feature of this framework is that the training process will use reinforcement learning. Within the given system, the encoder-decoder is considered a generator, whereas discriminators are reward modules. The discriminators and the generator are trained via maximum-likelihood and reinforcement learning, respectively, during training. This combined strategy ensures that the reward modules offer precise feedback, enhancing the quality of the generated reports. The reports created have been analysed using conventional language measures. Future work could improve these evaluation metrics by placing greater emphasis on clinical relevance and efficacy.

\section*{CRediT authorship contribution statement}
\textbf{Yogesh Kumar Meena}: Conceptualization, Methodology, Validation, Formal analysis, Writing - original draft, Writing - review \& editing, Supervision. \textbf{Saurabh Agarwal}: Methodology, Validation, Formal analysis, Writing - original draft, Writing - review \& editing, Visualization. \textbf{K. V. Arya}: Methodology, Writing - review \& editing.

\bibliographystyle{IEEEtran}
\bibliography{main}

\end{document}